%% file: iclr2024_conference.tex
\documentclass{article} 
\usepackage{iclr2024_conference,times}

\input{math_commands.tex}

\usepackage{hyperref}
\usepackage{url}
\usepackage{booktabs}
\usepackage{graphicx}
\usepackage{textgreek}
\usepackage{subcaption}
\usepackage{comment}
\usepackage{color}
\usepackage{tipa}
\usepackage[T4,T1]{fontenc}
\usepackage{xspace}
\newenvironment{tfour}{\fontencoding{T4}\selectfont}{}

\newcommand*{\yoruba}{Yor\`ub\'a\xspace}

\title{NaijaRC: A Multi-choice Reading Comprehension Dataset for Nigerian Languages}

\author{Anuoluwapo Aremu$^*$, Jesujoba Oluwadara Alabi$^{1*}$ , Daud Abolade$^{2*}$, Nkechinyere F. Aguobi$^2$  \\  
\textbf{Shamsuddeen Hassan Muhammad$^{3*}$, David Ifeoluwa Adelani$^{4*}$} \\
$^*$Masakhane NLP $^1$Saarland University $^2$University of Lagos $^3$Imperial College London\\
$^4$University College London \\
\texttt{aremuadeola97@gmail.com, d.adelani@ucl.ac.uk}
}

\iclrfinalcopy 
\begin{document}

\maketitle


\paragraph{Introduction}

Reading Comprehension (RC) requires the ability to read a text and demonstrate understanding by answering questions about it. This requires high-level reasoning, an understanding of natural language, and general real-world based knowledge for machines to deliver high-performance results. 
Developing AI models capable of understanding long paragraphs and answering questions is a major challenge, especially for under-resourced languages. 
With the quest to improve the machine performance of reading comprehension, there have been several efforts towards curation and acquisition of  datasets~\citep{Bajgar2016EmbracingDA,lai-etal-2017-race,kocisky-etal-2018-narrativeqa,xuan2023vnhsge}. 
However, many of these works are either in English or do not include any African languages. 

With a special lens on African languages, there has recently been a concentrated effort towards building datasets for NLP downstream tasks such as named entity recognition~\cite{eiselen-2016-government,masakhaner,adelani-etal-2022-masakhaner}, part-of-speech~\cite{nivre2016universal,Dione2023MasakhaPOSPT}, news topic classification~\cite{niyongabo-etal-2020-kinnews,hedderich-etal-2020-transfer,alabi-etal-2022-adapting, Adelani2023MasakhaNEWSNT}, and machine translation~\cite{reid-etal-2021-afromt,team2022NoLL,adelani-etal-2022-thousand}. However, little has been done on question answering and reading comprehension tasks. The closest dataset to ours is the AfriQA~\citep{ogundepo2023afriqa} dataset for open-retrieval question answering, with a primary focus on retrieving correct answers that are answerable on Wikipedia. Recently, Bebebele~\citep{bandarkar2023belebele}---a multi-choice RC dataset covering 122 languages was released, however, the questions and passages were translated from English dataset. To the best of our knowledge, there is no \textit{human-annotated} multi-choice reading comprehension dataset for African languages.  

In this paper, we create \textbf{NaijaRC}---a new multi-choice \textbf{N}igerian \textbf{R}eading \textbf{C}omprehension dataset that is based on high-school RC examination for three Nigerian national languages: Hausa (hau), Igbo (ibo), and \yoruba (yor). 
We provide baseline results by performing cross-lingual transfer using the Belebele training data which is majorly from RACE\footnote{RACE is based on English exams for middle and high school Chinese students, very similar to our dataset.} dataset~\citep{lai-etal-2017-race} based on several pre-trained encoder-only models. Additionally, we provide results by prompting large language models (LLMs) like GPT-4. We provide the code and data on GitHub~\footnote{\url{https://github.com/AremuAdeolaJr/NaijaRC}}. 

\paragraph{Nigerian languages covered}
The three Nigerian languages covered in the NaijaRC make use of Latin-based script and are tonal. Igbo and \yoruba additionally makes use of diacritics on their letters like \d{e}, \d{\`e}, \d{\'e}, and \d{u}. Hausa additional makes use of special letters like \texthtb, \texthtd, \texthtk, and \begin{tfour}\m{y}\end{tfour}. All the languages have more than 35 million native speakers according to Ethnologue~\citep{ethnologue} and are offered in high-school as a compulsory subject in the different regions where they are spoken: Hausa (North), Igbo (South-East), and \yoruba (South West / North Central). 

\paragraph{NaijaRC dataset collection}

We source past questions from the Hausa, Igbo, and \yoruba language subjects of the West African Senior School Certificate Examination (WASSCE). The subjects are optionally taken by final-year high school  students in south-western Nigeria. Prior to 2017, it used to be compulsory to take this subject. We only found printed versions of the past questions, thus, we had to make use of Optical Character Recognition tools like Google Lens and Apple's OCR to convert the printed format into a digitised version. Due to the diacritics used in the language, the OCR process has some quality issues, which we had to fix manually.


The West African Examination Council questions test many aspects of students' competence. However, with regard to the focus of our research, only the comprehension passages and questions were extracted. It is important to note that the number of questions generated for each passage varies from year to year. However, on average, there were five questions for each of the two passages annually.

Our dataset contains comprehension \textbf{passages}, \textbf{questions}, and \textbf{answers} which were carefully validated, verified, and cleaned by native speakers and linguists of the languages. We removed questions relating to the semantics of specific italics or underlined words in the context of usage, as most language models cannot handle them. 
\autoref{tab:results}(a) provides the data statistics and split for the different language. In Appendix, we also, provide an example of a passage, question and multi-choice answers for \yoruba and the translation to English. 

\begin{table}[t]
    \footnotesize
    \begin{subtable}{0.37\linewidth}
      \centering
        \begin{tabular}{lr|rrr}
        \toprule
        & \textbf{Belebele} & \multicolumn{3}{c}{\textbf{NaijaRC}}  \\
        & \textbf{eng} & \textbf{hau} & \textbf{ibo}  & \textbf{yor} \\
        \midrule
        TRAIN & 67,541 & - & - & 50 \\
        DEV & 3,773 & - & - & 25 \\
        TEST & -- & 88 & 88 & 171 \\
        \bottomrule
        \end{tabular}
        \vspace{0.4mm}
        \label{tab:data_split}
        \caption{Data split for Belebele and NaijaRC}
    \end{subtable} %
    \begin{subtable}{0.67\linewidth}
      \centering
      \setlength\tabcolsep{4pt} 
        \begin{tabular}{p{26mm}rrrr|r}
        \toprule
         & \multicolumn{5}{c}{\textbf{Accuracy}}\\
        \textbf{Model} & \textbf{eng} & \textbf{hau} & \textbf{ibo}  & \textbf{yor} & \textbf{AVG}\\
        \midrule
            AfroXLMR-base & 67.3 & 43.2 & 42.0 & 34.0 & 46.6 \\
            Serengeti & 62.5 & 50.0 & 44.3 & 33.0 & 47.4 \\
            OFA-768 & 64.5 & 43.2 & 48.9 & 28.3 & 46.2 \\
            GPT-3.5 Turbo & 66.3 & 35.2 & 30.7 & 27.7 & 40.0 \\
            GPT-4 Turbo & 82.9 & 48.9 & 40.9 & 33.0 & 51.4 \\
        
        \bottomrule
        \end{tabular}
        \label{tab:baseline}
        \caption{Model accuracy. PLMs fine-tuned on Belebele's train split. }
    \end{subtable}
    \vspace{-5mm}
    \caption{\textbf{Data split and benchmark results} for Belebele (on DEV set) and NaijaRC (on test set).}
    \label{tab:results}
\end{table}

\paragraph{Experimental setup} In this work, we evaluated the performance of fine-tuned pre-trained encoder-only models, specifically AfroXLMR-base~\cite{alabi-etal-2022-adapting}, Serengeti~\citep{adebara-etal-2023-serengeti} and OFA-768~\citep{liu2023ofa},  as well as two large language models (LLMs), namely GPT-3.5 and GPT-4. For the encoder based PLMs, we evaluated their performance in a zero-shot cross-lingual setting where we fine-tuned the model on the training split of Belebele~\footnote{the training data is a combination of several multiple choice datasets and can be retrieved here https://github.com/facebookresearch/belebele}~\citep{bandarkar2023belebele} dataset for 3 epochs and evaluated the performance of the resulting model on NaijaRC. Furthermore, we prompted GPT-3.5 and GPT-4 providing them with the question, options, and context paragraph. The output of these models are then mapped back to the option labels for evaluation. All these models are evaluated using accuracy as the metric.

\paragraph{Experimental results} The experimental results are presented in \autoref{tab:results}(b) including the models' performance on the Belebele's development set and the zero-shot transfer performance on NaijaRC test sets. The result shows that GPT-4 achieved the highest overall accuracy across the four languages, with an accuracy of $51.4\%$. This score can be attributed to its exceptional performance on the Belebele development set, where it achieved an accuracy of $82.9\%$. While there is a possibility that the model benefited from pretraining on the exact data in the development, it is crucial to note that this is not a scenario of "winner takes it all".  The zero-shot cross-lingual transfer result with PLMs show that different models performed best in specific languages. AfroXLMR-base has highest accuracy on \yoruba ($34.0\%$), OFA-768 outperformed others on Igbo ($48.9\%$), and Serengeti outperformed others on Hausa ($50.0\%$). Therefore, when excluding English from the average calculation, Serengeti emerged with the highest overall performance. However, GPT-3.5 and GPT-4 showed relatively lower performance compared to multilingual PLMs on the Nigerian languages, with GPT-4 emerging as the superior performer between the two.


\paragraph{Conclusion and Future work}
In this paper, we presented \textbf{NaijaRC}---a new reading comprehension dataset for Nigerian languages. We compared the zero-shot cross-lingual transfer with three PLMs: AfroXLMR-base, Serengeti, and OFA-768 from an English dataset. We also prompted GPT-3.5 and GPT.4, and showed that these models obtained lower performance on NaijaRC dataset when compared to the PLMs.  

As future work, we plan to extend our evaluation to few-shot settings. Specifically, we will explore different approaches that can utilize a few examples, like 50 (passage, question, answer)-tuples for the effective adaptation of existing reading comprehension models. 

\section*{Acknowledgment}
David Adelani acknowledges the support of DeepMind Academic Fellowship programme. We are grateful to OpenAI for providing API credits through their Researcher Access API programme to Masakhane for the evaluation of GPT-3.5 and GPT-4 large language models. Jesujoba Alabi was partially supported by the BMBF’s (German Federal Ministry of Education and Research) SLIK project under the grant 01IS22015C.

\bibliography{iclr2024_conference}
\bibliographystyle{iclr2024_conference}

\appendix
\section{Appendix}
Example of the passage in \yoruba

\begin{figure}[t]
  \centering
  \includegraphics[width=\columnwidth]{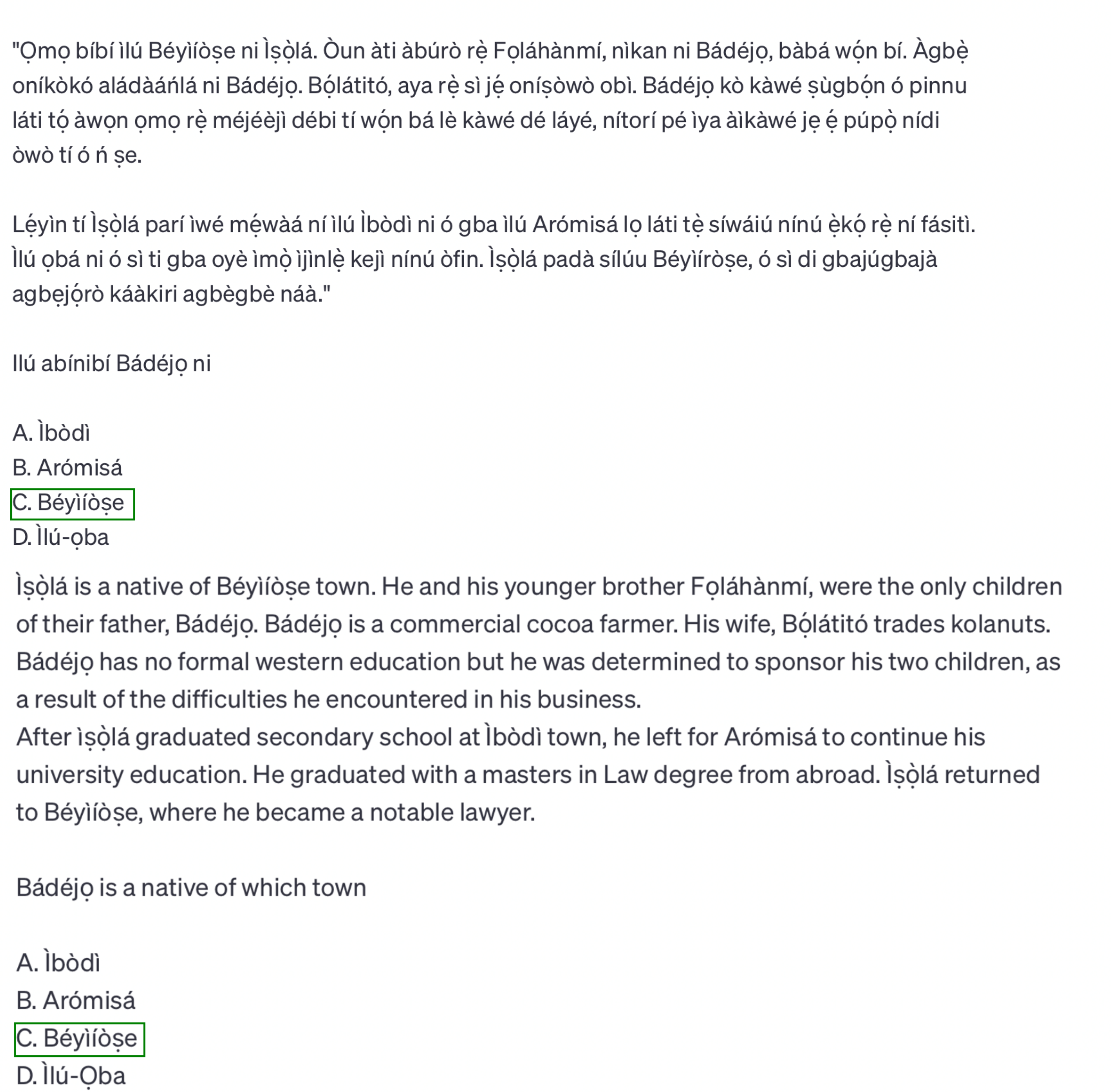}
  \caption{An example of a passage in \yoruba, a question and corresponding options (A-D). Where C is the correct option. We provide an expert translation in English.}
  \vspace{-3mm}
  \label{img:sample}
\end{figure}

\end{document}

%% file: math_commands.tex
\usepackage{amsmath,amsfonts,bm}

\def\eqref#1{equation~\ref{#1}}

\def\1{\bm{1}}

\DeclareMathAlphabet{\mathsfit}{\encodingdefault}{\sfdefault}{m}{sl}
\SetMathAlphabet{\mathsfit}{bold}{\encodingdefault}{\sfdefault}{bx}{n}













%% file: iclr2024_conference.bbl
\begin{thebibliography}{21}
\providecommand{\natexlab}[1]{#1}
\providecommand{\url}[1]{\texttt{#1}}
\expandafter\ifx\csname urlstyle\endcsname\relax
  \providecommand{\doi}[1]{doi: #1}\else
  \providecommand{\doi}{doi: \begingroup \urlstyle{rm}\Url}\fi

\bibitem[Adebara et~al.(2023)Adebara, Elmadany, Abdul-Mageed, and Alcoba~Inciarte]{adebara-etal-2023-serengeti}
Ife Adebara, AbdelRahim Elmadany, Muhammad Abdul-Mageed, and Alcides Alcoba~Inciarte.
\newblock {SERENGETI}: Massively multilingual language models for {A}frica.
\newblock In Anna Rogers, Jordan Boyd-Graber, and Naoaki Okazaki (eds.), \emph{Findings of the Association for Computational Linguistics: ACL 2023}, pp.\  1498--1537, Toronto, Canada, July 2023. Association for Computational Linguistics.
\newblock \doi{10.18653/v1/2023.findings-acl.97}.
\newblock URL \url{https://aclanthology.org/2023.findings-acl.97}.

\bibitem[Adelani et~al.(2022{\natexlab{a}})Adelani, Alabi, Fan, Kreutzer, Shen, Reid, Ruiter, Klakow, Nabende, Chang, Gwadabe, Sackey, Dossou, Emezue, Leong, Beukman, Muhammad, Jarso, Yousuf, Niyongabo~Rubungo, Hacheme, Wairagala, Nasir, Ajibade, Ajayi, Gitau, Abbott, Ahmed, Ochieng, Aremu, Ogayo, Mukiibi, Ouoba~Kabore, Kalipe, Mbaye, Tapo, Memdjokam~Koagne, Munkoh-Buabeng, Wagner, Abdulmumin, Awokoya, Buzaaba, Sibanda, Bukula, and Manthalu]{adelani-etal-2022-thousand}
David Adelani, Jesujoba Alabi, Angela Fan, Julia Kreutzer, Xiaoyu Shen, Machel Reid, Dana Ruiter, Dietrich Klakow, Peter Nabende, Ernie Chang, Tajuddeen Gwadabe, Freshia Sackey, Bonaventure F.~P. Dossou, Chris Emezue, Colin Leong, Michael Beukman, Shamsuddeen Muhammad, Guyo Jarso, Oreen Yousuf, Andre Niyongabo~Rubungo, Gilles Hacheme, Eric~Peter Wairagala, Muhammad~Umair Nasir, Benjamin Ajibade, Tunde Ajayi, Yvonne Gitau, Jade Abbott, Mohamed Ahmed, Millicent Ochieng, Anuoluwapo Aremu, Perez Ogayo, Jonathan Mukiibi, Fatoumata Ouoba~Kabore, Godson Kalipe, Derguene Mbaye, Allahsera~Auguste Tapo, Victoire Memdjokam~Koagne, Edwin Munkoh-Buabeng, Valencia Wagner, Idris Abdulmumin, Ayodele Awokoya, Happy Buzaaba, Blessing Sibanda, Andiswa Bukula, and Sam Manthalu.
\newblock A few thousand translations go a long way! leveraging pre-trained models for {A}frican news translation.
\newblock In Marine Carpuat, Marie-Catherine de~Marneffe, and Ivan~Vladimir Meza~Ruiz (eds.), \emph{Proceedings of the 2022 Conference of the North American Chapter of the Association for Computational Linguistics: Human Language Technologies}, pp.\  3053--3070, Seattle, United States, July 2022{\natexlab{a}}. Association for Computational Linguistics.
\newblock \doi{10.18653/v1/2022.naacl-main.223}.
\newblock URL \url{https://aclanthology.org/2022.naacl-main.223}.

\bibitem[Adelani et~al.(2022{\natexlab{b}})Adelani, Neubig, Ruder, Rijhwani, Beukman, Palen-Michel, Lignos, Alabi, Muhammad, Nabende, Dione, Bukula, Mabuya, Dossou, Sibanda, Buzaaba, Mukiibi, Kalipe, Mbaye, Taylor, Kabore, Emezue, Aremu, Ogayo, Gitau, Munkoh-Buabeng, Memdjokam~Koagne, Tapo, Macucwa, Marivate, Elvis, Gwadabe, Adewumi, Ahia, Nakatumba-Nabende, Mokono, Ezeani, Chukwuneke, Oluwaseun~Adeyemi, Hacheme, Abdulmumin, Ogundepo, Yousuf, Moteu, and Klakow]{adelani-etal-2022-masakhaner}
David Adelani, Graham Neubig, Sebastian Ruder, Shruti Rijhwani, Michael Beukman, Chester Palen-Michel, Constantine Lignos, Jesujoba Alabi, Shamsuddeen Muhammad, Peter Nabende, Cheikh M.~Bamba Dione, Andiswa Bukula, Rooweither Mabuya, Bonaventure F.~P. Dossou, Blessing Sibanda, Happy Buzaaba, Jonathan Mukiibi, Godson Kalipe, Derguene Mbaye, Amelia Taylor, Fatoumata Kabore, Chris~Chinenye Emezue, Anuoluwapo Aremu, Perez Ogayo, Catherine Gitau, Edwin Munkoh-Buabeng, Victoire Memdjokam~Koagne, Allahsera~Auguste Tapo, Tebogo Macucwa, Vukosi Marivate, Mboning~Tchiaze Elvis, Tajuddeen Gwadabe, Tosin Adewumi, Orevaoghene Ahia, Joyce Nakatumba-Nabende, Neo~Lerato Mokono, Ignatius Ezeani, Chiamaka Chukwuneke, Mofetoluwa Oluwaseun~Adeyemi, Gilles~Quentin Hacheme, Idris Abdulmumin, Odunayo Ogundepo, Oreen Yousuf, Tatiana Moteu, and Dietrich Klakow.
\newblock {M}asakha{NER} 2.0: {A}frica-centric transfer learning for named entity recognition.
\newblock In \emph{Proceedings of the 2022 Conference on Empirical Methods in Natural Language Processing}, pp.\  4488--4508, Abu Dhabi, United Arab Emirates, December 2022{\natexlab{b}}. Association for Computational Linguistics.
\newblock URL \url{https://aclanthology.org/2022.emnlp-main.298}.

\bibitem[Adelani et~al.(2021)Adelani, Abbott, Neubig, D’souza, Kreutzer, Lignos, Palen-Michel, Buzaaba, Rijhwani, Ruder, Mayhew, Azime, Muhammad, Emezue, Nakatumba-Nabende, Ogayo, Anuoluwapo, Gitau, Mbaye, Alabi, Yimam, Gwadabe, Ezeani, Niyongabo, Mukiibi, Otiende, Orife, David, Ngom, Adewumi, Rayson, Adeyemi, Muriuki, Anebi, Chukwuneke, Odu, Wairagala, Oyerinde, Siro, Bateesa, Oloyede, Wambui, Akinode, Nabagereka, Katusiime, Awokoya, MBOUP, Gebreyohannes, Tilaye, Nwaike, Wolde, Faye, Sibanda, Ahia, Dossou, Ogueji, DIOP, Diallo, Akinfaderin, Marengereke, and Osei]{masakhaner}
David~Ifeoluwa Adelani, Jade Abbott, Graham Neubig, Daniel D’souza, Julia Kreutzer, Constantine Lignos, Chester Palen-Michel, Happy Buzaaba, Shruti Rijhwani, Sebastian Ruder, Stephen Mayhew, Israel~Abebe Azime, Shamsuddeen~H. Muhammad, Chris~Chinenye Emezue, Joyce Nakatumba-Nabende, Perez Ogayo, Aremu Anuoluwapo, Catherine Gitau, Derguene Mbaye, Jesujoba Alabi, Seid~Muhie Yimam, Tajuddeen~Rabiu Gwadabe, Ignatius Ezeani, Rubungo~Andre Niyongabo, Jonathan Mukiibi, Verrah Otiende, Iroro Orife, Davis David, Samba Ngom, Tosin Adewumi, Paul Rayson, Mofetoluwa Adeyemi, Gerald Muriuki, Emmanuel Anebi, Chiamaka Chukwuneke, Nkiruka Odu, Eric~Peter Wairagala, Samuel Oyerinde, Clemencia Siro, Tobius~Saul Bateesa, Temilola Oloyede, Yvonne Wambui, Victor Akinode, Deborah Nabagereka, Maurice Katusiime, Ayodele Awokoya, Mouhamadane MBOUP, Dibora Gebreyohannes, Henok Tilaye, Kelechi Nwaike, Degaga Wolde, Abdoulaye Faye, Blessing Sibanda, Orevaoghene Ahia, Bonaventure F.~P. Dossou, Kelechi Ogueji, Thierno~Ibrahima DIOP,
  Abdoulaye Diallo, Adewale Akinfaderin, Tendai Marengereke, and Salomey Osei.
\newblock {MasakhaNER: Named Entity Recognition for African Languages}.
\newblock \emph{Transactions of the Association for Computational Linguistics}, 9:\penalty0 1116--1131, 10 2021.
\newblock ISSN 2307-387X.
\newblock \doi{10.1162/tacl_a_00416}.
\newblock URL \url{https://doi.org/10.1162/tacl\_a\_00416}.

\bibitem[Adelani et~al.(2023)Adelani, Masiak, Azime, Alabi, Tonja, Mwase, Ogundepo, Dossou, Oladipo, Nixdorf, Emezue, Al-Azzawi, Sibanda, David, Ndolela, Mukiibi, Ajayi, Ngoli, Odhiambo, Owodunni, Obiefuna, Muhammad, Abdullahi, Yigezu, Gwadabe, Abdulmumin, Bame, Awoyomi, Shode, Adelani, Kailani, Omotayo, Adeeko, Abeeb, Aremu, Samuel, Siro, Kimotho, Ogbu, Mbonu, Chukwuneke, Fanijo, Ojo, Awosan, Guge, Sari, Nyatsine, Sidume, Yousuf, Oduwole, Kimanuka, Tshinu, Diko, Nxakama, Johar, Gebre, Mohamed, Mohamed, Hassan, Mehamed, Ngabire, and Stenetorp]{Adelani2023MasakhaNEWSNT}
David~Ifeoluwa Adelani, Marek Masiak, Israel~Abebe Azime, Jesujoba~Oluwadara Alabi, Atnafu~Lambebo Tonja, Christine Mwase, Odunayo Ogundepo, Bonaventure F.~P. Dossou, Akintunde Oladipo, Doreen Nixdorf, Chris~Chinenye Emezue, Sana Al-Azzawi, Blessing~K. Sibanda, Davis David, Lolwethu Ndolela, Jonathan Mukiibi, Tunde~Oluwaseyi Ajayi, Tatiana~Moteu Ngoli, Brian Odhiambo, Abraham~Toluwase Owodunni, Nnaemeka~C. Obiefuna, Shamsuddeen~Hassan Muhammad, Saheed~Salahudeen Abdullahi, Mesay~Gemeda Yigezu, Tajuddeen~Rabiu Gwadabe, Idris Abdulmumin, Mahlet~Taye Bame, Oluwabusayo~Olufunke Awoyomi, Iyanuoluwa Shode, Tolulope~Anu Adelani, Habiba~Abdulganiy Kailani, Abdul-Hakeem Omotayo, Adetola Adeeko, Afolabi Abeeb, Anuoluwapo Aremu, Olanrewaju Samuel, Clemencia Siro, Wangari Kimotho, Onyekachi~Raphael Ogbu, Chinedu~E. Mbonu, Chiamaka~I. Chukwuneke, Samuel Fanijo, Jessica Ojo, Oyinkansola~F. Awosan, Tadesse~Kebede Guge, Sakayo~Toadoum Sari, Pamela Nyatsine, Freedmore Sidume, Oreen Yousuf, Mardiyyah Oduwole, Ussen~Abre
  Kimanuka, Kanda~Patrick Tshinu, Thina Diko, Siyanda Nxakama, Abdulmejid~Tuni Johar, Sinodos Gebre, Muhidin Mohamed, S.~A. Mohamed, Fuad~Mire Hassan, Moges~Ahmed Mehamed, Evrard Ngabire, and Pontus Stenetorp.
\newblock Masakhanews: News topic classification for african languages.
\newblock 2023.

\bibitem[Alabi et~al.(2022)Alabi, Adelani, Mosbach, and Klakow]{alabi-etal-2022-adapting}
Jesujoba~O. Alabi, David~Ifeoluwa Adelani, Marius Mosbach, and Dietrich Klakow.
\newblock Adapting pre-trained language models to {A}frican languages via multilingual adaptive fine-tuning.
\newblock In \emph{Proceedings of the 29th International Conference on Computational Linguistics}, pp.\  4336--4349, Gyeongju, Republic of Korea, October 2022. International Committee on Computational Linguistics.
\newblock URL \url{https://aclanthology.org/2022.coling-1.382}.

\bibitem[Bajgar et~al.(2016)Bajgar, Kadlec, and Kleindienst]{Bajgar2016EmbracingDA}
Ondrej Bajgar, Rudolf Kadlec, and Jan Kleindienst.
\newblock Embracing data abundance: Booktest dataset for reading comprehension.
\newblock \emph{ArXiv}, abs/1610.00956, 2016.

\bibitem[Bandarkar et~al.(2023)Bandarkar, Liang, Muller, Artetxe, Shukla, Husa, Goyal, Krishnan, Zettlemoyer, and Khabsa]{bandarkar2023belebele}
Lucas Bandarkar, Davis Liang, Benjamin Muller, Mikel Artetxe, Satya~Narayan Shukla, Donald Husa, Naman Goyal, Abhinandan Krishnan, Luke Zettlemoyer, and Madian Khabsa.
\newblock The belebele benchmark: a parallel reading comprehension dataset in 122 language variants.
\newblock \emph{arXiv preprint arXiv:2308.16884}, 2023.

\bibitem[Dione et~al.(2023)Dione, Adelani, Nabende, Alabi, Sindane, Buzaaba, Muhammad, Emezue, Ogayo, Aremu, Gitau, Mbaye, Mukiibi, Sibanda, Dossou, Bukula, Mabuya, Tapo, Munkoh-Buabeng, victoire Memdjokam~Koagne, Kabore, Taylor, Kalipe, Macucwa, Marivate, Gwadabe, Elvis, Onyenwe, Atindogbe, Adelani, Akinade, Samuel, Nahimana, Musabeyezu, Niyomutabazi, Chimhenga, Gotosa, Mizha, Agbolo, Traore, Uchechukwu, Yusuf, Abdullahi, and Klakow]{Dione2023MasakhaPOSPT}
Cheikh M.~Bamba Dione, David Adelani, Peter Nabende, Jesujoba Alabi, Thapelo Sindane, Happy Buzaaba, Shamsuddeen~Hassan Muhammad, Chris~Chinenye Emezue, Perez Ogayo, Anuoluwapo Aremu, Catherine Gitau, Derguene Mbaye, Jonathan Mukiibi, Blessing Sibanda, Bonaventure F.~P. Dossou, Andiswa Bukula, Rooweither Mabuya, Allahsera~Auguste Tapo, Edwin Munkoh-Buabeng, victoire Memdjokam~Koagne, Fatoumata~Ouoba Kabore, Amelia Taylor, Godson Kalipe, Tebogo Macucwa, Vukosi Marivate, Tajuddeen Gwadabe, Mboning~Tchiaze Elvis, Ikechukwu Onyenwe, Gratien Atindogbe, Tolulope Adelani, Idris Akinade, Olanrewaju Samuel, Marien Nahimana, Th'eogene Musabeyezu, Emile Niyomutabazi, Ester Chimhenga, Kudzai Gotosa, Patrick Mizha, Apelete Agbolo, Seydou Traore, Chinedu Uchechukwu, Aliyu Yusuf, Muhammad Abdullahi, and Dietrich Klakow.
\newblock Masakhapos: Part-of-speech tagging for typologically diverse african languages.
\newblock 2023.

\bibitem[Eberhard et~al.(2020)Eberhard, Simons, and (eds.)]{ethnologue}
David~M. Eberhard, Gary~F. Simons, and Charles D.~Fennig (eds.).
\newblock Ethnologue: Languages of the world. twenty-third edition., 2020.
\newblock URL \url{http://www.ethnologue.com}.

\bibitem[Eiselen(2016)]{eiselen-2016-government}
Roald Eiselen.
\newblock Government domain named entity recognition for {S}outh {A}frican languages.
\newblock In Nicoletta Calzolari, Khalid Choukri, Thierry Declerck, Sara Goggi, Marko Grobelnik, Bente Maegaard, Joseph Mariani, Helene Mazo, Asuncion Moreno, Jan Odijk, and Stelios Piperidis (eds.), \emph{Proceedings of the Tenth International Conference on Language Resources and Evaluation ({LREC}'16)}, pp.\  3344--3348, Portoro{\v{z}}, Slovenia, May 2016. European Language Resources Association (ELRA).
\newblock URL \url{https://aclanthology.org/L16-1533}.

\bibitem[Hedderich et~al.(2020)Hedderich, Adelani, Zhu, Alabi, Markus, and Klakow]{hedderich-etal-2020-transfer}
Michael~A. Hedderich, David Adelani, Dawei Zhu, Jesujoba Alabi, Udia Markus, and Dietrich Klakow.
\newblock Transfer learning and distant supervision for multilingual transformer models: A study on {A}frican languages.
\newblock In Bonnie Webber, Trevor Cohn, Yulan He, and Yang Liu (eds.), \emph{Proceedings of the 2020 Conference on Empirical Methods in Natural Language Processing (EMNLP)}, pp.\  2580--2591, Online, November 2020. Association for Computational Linguistics.
\newblock \doi{10.18653/v1/2020.emnlp-main.204}.
\newblock URL \url{https://aclanthology.org/2020.emnlp-main.204}.

\bibitem[Ko{\v{c}}isk{\'y} et~al.(2018)Ko{\v{c}}isk{\'y}, Schwarz, Blunsom, Dyer, Hermann, Melis, and Grefenstette]{kocisky-etal-2018-narrativeqa}
Tom{\'a}{\v{s}} Ko{\v{c}}isk{\'y}, Jonathan Schwarz, Phil Blunsom, Chris Dyer, Karl~Moritz Hermann, G{\'a}bor Melis, and Edward Grefenstette.
\newblock The {N}arrative{QA} reading comprehension challenge.
\newblock \emph{Transactions of the Association for Computational Linguistics}, 6:\penalty0 317--328, 2018.
\newblock \doi{10.1162/tacl_a_00023}.
\newblock URL \url{https://aclanthology.org/Q18-1023}.

\bibitem[Lai et~al.(2017)Lai, Xie, Liu, Yang, and Hovy]{lai-etal-2017-race}
Guokun Lai, Qizhe Xie, Hanxiao Liu, Yiming Yang, and Eduard Hovy.
\newblock {RACE}: Large-scale {R}e{A}ding comprehension dataset from examinations.
\newblock In Martha Palmer, Rebecca Hwa, and Sebastian Riedel (eds.), \emph{Proceedings of the 2017 Conference on Empirical Methods in Natural Language Processing}, pp.\  785--794, Copenhagen, Denmark, September 2017. Association for Computational Linguistics.
\newblock \doi{10.18653/v1/D17-1082}.
\newblock URL \url{https://aclanthology.org/D17-1082}.

\bibitem[Liu et~al.(2023)Liu, Lin, Wang, and Schütze]{liu2023ofa}
Yihong Liu, Peiqin Lin, Mingyang Wang, and Hinrich Schütze.
\newblock Ofa: A framework of initializing unseen subword embeddings for efficient large-scale multilingual continued pretraining, 2023.

\bibitem[Nivre et~al.(2016)Nivre, De~Marneffe, Ginter, Goldberg, Hajic, Manning, McDonald, Petrov, Pyysalo, Silveira, et~al.]{nivre2016universal}
Joakim Nivre, Marie-Catherine De~Marneffe, Filip Ginter, Yoav Goldberg, Jan Hajic, Christopher~D Manning, Ryan McDonald, Slav Petrov, Sampo Pyysalo, Natalia Silveira, et~al.
\newblock Universal dependencies v1: A multilingual treebank collection.
\newblock In \emph{Proceedings of the Tenth International Conference on Language Resources and Evaluation (LREC'16)}, pp.\  1659--1666, 2016.

\bibitem[Niyongabo et~al.(2020)Niyongabo, Hong, Kreutzer, and Huang]{niyongabo-etal-2020-kinnews}
Rubungo~Andre Niyongabo, Qu~Hong, Julia Kreutzer, and Li~Huang.
\newblock {KINNEWS} and {KIRNEWS}: Benchmarking cross-lingual text classification for {K}inyarwanda and {K}irundi.
\newblock In Donia Scott, Nuria Bel, and Chengqing Zong (eds.), \emph{Proceedings of the 28th International Conference on Computational Linguistics}, pp.\  5507--5521, Barcelona, Spain (Online), December 2020. International Committee on Computational Linguistics.
\newblock \doi{10.18653/v1/2020.coling-main.480}.
\newblock URL \url{https://aclanthology.org/2020.coling-main.480}.

\bibitem[NLLB-Team et~al.(2022)NLLB-Team, Costa-juss{\`a}, Cross, cCelebi, Elbayad, Heafield, Heffernan, Kalbassi, Lam, Licht, Maillard, Sun, Wang, Wenzek, Youngblood, Akula, Barrault, Gonzalez, Hansanti, Hoffman, Jarrett, Sadagopan, Rowe, Spruit, Tran, Andrews, Ayan, Bhosale, Edunov, Fan, Gao, Goswami, Guzm'an, Koehn, Mourachko, Ropers, Saleem, Schwenk, and Wang]{team2022NoLL}
NLLB-Team, Marta~Ruiz Costa-juss{\`a}, James Cross, Onur cCelebi, Maha Elbayad, Kenneth Heafield, Kevin Heffernan, Elahe Kalbassi, Janice Lam, Daniel Licht, Jean Maillard, Anna Sun, Skyler Wang, Guillaume Wenzek, Alison Youngblood, Bapi Akula, Lo{\"i}c Barrault, Gabriel~Mejia Gonzalez, Prangthip Hansanti, John Hoffman, Semarley Jarrett, Kaushik~Ram Sadagopan, Dirk Rowe, Shannon~L. Spruit, C.~Tran, Pierre Andrews, Necip~Fazil Ayan, Shruti Bhosale, Sergey Edunov, Angela Fan, Cynthia Gao, Vedanuj Goswami, Francisco Guzm'an, Philipp Koehn, Alexandre Mourachko, Christophe Ropers, Safiyyah Saleem, Holger Schwenk, and Jeff Wang.
\newblock No language left behind: Scaling human-centered machine translation.
\newblock \emph{ArXiv}, abs/2207.04672, 2022.

\bibitem[Ogundepo et~al.(2023)Ogundepo, Gwadabe, Rivera, Clark, Ruder, Adelani, Dossou, DIOP, Sikasote, Hacheme, Buzaaba, Ezeani, Mabuya, Osei, Emezue, Kahira, Muhammad, Oladipo, Owodunni, Tonja, Shode, Asai, Ajayi, Siro, Arthur, Adeyemi, Ahia, Anuoluwapo, Awosan, Chukwuneke, Opoku, Ayodele, Otiende, Mwase, Sinkala, Rubungo, Ajisafe, Onwuegbuzia, Mbow, Niyomutabazi, Mukonde, Lawan, Ahmad, Alabi, Namukombo, Chinedu, Phiri, Putini, Mngoma, Amuok, Iro, and Adhiambo]{ogundepo2023afriqa}
Odunayo Ogundepo, Tajuddeen~R. Gwadabe, Clara~E. Rivera, Jonathan~H. Clark, Sebastian Ruder, David~Ifeoluwa Adelani, Bonaventure F.~P. Dossou, Abdou~Aziz DIOP, Claytone Sikasote, Gilles Hacheme, Happy Buzaaba, Ignatius Ezeani, Rooweither Mabuya, Salomey Osei, Chris Emezue, Albert~Njoroge Kahira, Shamsuddeen~H. Muhammad, Akintunde Oladipo, Abraham~Toluwase Owodunni, Atnafu~Lambebo Tonja, Iyanuoluwa Shode, Akari Asai, Tunde~Oluwaseyi Ajayi, Clemencia Siro, Steven Arthur, Mofetoluwa Adeyemi, Orevaoghene Ahia, Aremu Anuoluwapo, Oyinkansola Awosan, Chiamaka Chukwuneke, Bernard Opoku, Awokoya Ayodele, Verrah Otiende, Christine Mwase, Boyd Sinkala, Andre~Niyongabo Rubungo, Daniel~A. Ajisafe, Emeka~Felix Onwuegbuzia, Habib Mbow, Emile Niyomutabazi, Eunice Mukonde, Falalu~Ibrahim Lawan, Ibrahim~Said Ahmad, Jesujoba~O. Alabi, Martin Namukombo, Mbonu Chinedu, Mofya Phiri, Neo Putini, Ndumiso Mngoma, Priscilla~A. Amuok, Ruqayya~Nasir Iro, and Sonia Adhiambo.
\newblock Afriqa: Cross-lingual open-retrieval question answering for african languages, 2023.

\bibitem[Reid et~al.(2021)Reid, Hu, Neubig, and Matsuo]{reid-etal-2021-afromt}
Machel Reid, Junjie Hu, Graham Neubig, and Yutaka Matsuo.
\newblock {A}fro{MT}: Pretraining strategies and reproducible benchmarks for translation of 8 {A}frican languages.
\newblock In Marie-Francine Moens, Xuanjing Huang, Lucia Specia, and Scott Wen-tau Yih (eds.), \emph{Proceedings of the 2021 Conference on Empirical Methods in Natural Language Processing}, pp.\  1306--1320, Online and Punta Cana, Dominican Republic, November 2021. Association for Computational Linguistics.
\newblock \doi{10.18653/v1/2021.emnlp-main.99}.
\newblock URL \url{https://aclanthology.org/2021.emnlp-main.99}.

\bibitem[Xuan-Quy et~al.(2023)Xuan-Quy, Ngoc-Bich, The-Duy, Xuan-Dung, Bac-Bien, Van-Tien, Thi-My-Thanh, and Hong-Phuoc]{xuan2023vnhsge}
Dao Xuan-Quy, Le~Ngoc-Bich, Vo~The-Duy, Phan Xuan-Dung, Ngo Bac-Bien, Nguyen Van-Tien, Nguyen Thi-My-Thanh, and Nguyen Hong-Phuoc.
\newblock Vnhsge: Vietnamese high school graduation examination dataset for large language models.
\newblock \emph{arXiv preprint arXiv:2305.12199}, 2023.

\end{thebibliography}
